\documentclass[a4paper,fleqn]{cas-dc}

\usepackage[numbers]{natbib}
\usepackage{algorithm}
\usepackage{algcompatible}

\def\tsc#1{\csdef{#1}{\textsc{\lowercase{#1}}\xspace}}
\tsc{WGM}
\tsc{QE}
\begin{document}
\let\WriteBookmarks\relax
\def\floatpagepagefraction{1}
\def\textpagefraction{.001}

\shorttitle{}    

\shortauthors{Cao et al.}  

\title [mode = title]{IAENet: An Importance-Aware Ensemble Model for 3D Point Cloud-Based Anomaly Detection}  




\author[1]{Xuanming Cao}%[<options>]

% \cormark[1]


\ead{xcao743@connect.hkust-gz.edu.cn}


\credit{Writing – original draft, Conceptualization, Methodology, Formal analysis, Validation, Software, Visualization}

\affiliation[1]{organization={Smart Manufacturing Thrust},
            addressline={The Hong Kong University of Science and 
            Technology (Guangzhou)}, 
            city={Guangzhou},
%          citysep={}, % Uncomment if no comma needed between city and postcode
            postcode={511453}, 
            country={China}}

\author[2]{Chengyu Tao}%[]

% Footnote of the second author

% Email id of the second author
\ead{ctaoaa@connect.ust.hk}


\credit{Writing – review and editing, Data curation, Formal analysis}

\affiliation[2]{organization={Interdisciplinary Programs Office},
            addressline={The Hong Kong University of Science and 
            Technology}, 
            city={Hong Kong SAR},
%          citysep={}, % Uncomment if no comma needed between city and postcode
            postcode={999077}, 
            country={China}}

\author[1]{Yifeng Cheng}%[]
\ead{yifengcheng@hkust-gz.edu.cn}
\credit{Writing – review and editing, Investigation}

\author[1,2,3]{Juan Du}[orcid=0000-0002-6018-2972]
\cormark[1]
\ead{juandu@ust.hk}
\affiliation[3]{organization={Department of Mechanical and Aerospace Engineering},
            addressline={The Hong Kong University of Science and 
            Technology}, 
            city={Hong Kong SAR},
%          citysep={}, % Uncomment if no comma needed between city and postcode
            postcode={999077}, 
            country={China}}
\credit{Writing – review and editing, Project administration, Funding acquisition, Supervision}

\cortext[1]{Corresponding author}



\begin{abstract}
Surface anomaly detection is pivotal for ensuring product quality in industrial manufacturing. While 2D image-based methods have achieved remarkable success, 3D point cloud-based detection remains underexplored despite its richer geometric cues. We argue that the key bottleneck is the absence of powerful pretrained foundation backbones in 3D comparable to those in 2D. To bridge this gap, we propose Importance-Aware Ensemble Network (IAENet), an ensemble framework that synergizes 2D pretrained expert with 3D expert models. However, naively fusing predictions from disparate sources is non-trivial: existing strategies can be affected by a poorly performing modality and thus degrade overall accuracy. To address this challenge, We introduce an novel Importance-Aware Fusion (IAF) module that dynamically assesses the contribution of each source and reweights their anomaly scores. Furthermore, we devise critical loss functions that explicitly guide the optimization of IAF, enabling it to combine the collective knowledge of the source experts but also preserve their unique strengths, thereby enhancing the overall performance of anomaly detection. Extensive experiments on MVTec 3D-AD demonstrate that our IAENet achieves a new state-of-the-art with a markedly lower false positive rate, underscoring its practical value for industrial deployment.
\end{abstract}


% \begin{highlights}
% \item IAENet: 2D pretrained + 3D expert ensemble for 3D point cloud based-anomaly detection.
% \item Importance-Aware Fusion adaptively weights individual model predictions.
% \item Novel loss ensures fusion outcomes are unaffected by inferior model predictions.
% \item Promising results on MVTec 3D-AD with industry-friendly low false positive rate.
% \end{highlights}

%\nocite{*}

% Keywords

\maketitle

\section{Introduction}
\label{sec:intro}

Surface anomaly detection plays an essential role in ensuring that products meet standards and specifications across a multitude of industrial applications \cite{cao20253d}. Traditionally, this task has been heavily dependent on manual visual inspection \cite{du20253d}, which is not only labor-intensive but also highly subjective and inefficient. With the advancement of machine vision technology, the possibility of replacing manual inspection with automated methods has become increasingly feasible, offering higher efficiency, accuracy, and robustness \cite{lin2025survey}.

Although well-established 2D image-based detection methods have yielded promising results \cite{roth2022towards,defard2021padim}, they are not without limitations. These methods are often influenced by lighting conditions and may struggle to identify geometric anomalies that are similar in color to the background. Additionally, they lack the capacity to discern geometric details such as size and shape \cite{cao20253d}. The advance of 3D scanning technology has made the acquisition of high-quality 3D point cloud data more accessible and affordable, leading to a surge in research focused on 3D point cloud-based anomaly detection.

Due to the scarcity of anomalous samples in industrial settings, most anomaly detection methods are unsupervised, focusing on learning the distribution of normal samples. For instance, Roth et al. \cite{roth2022towards} introduce a memory bank approach where test sample features are compared against stored normal features to identify anomalies. Another method \cite{bergmann2023anomaly} employs a teacher-student architecture to generate anomaly scores based on feature discrepancies between a pretrained teacher model and a student model trained on normal data.
In addition to these unimodal methods, recent studies have explored the fusion of image and point cloud data. A representative example \cite{horwitz2023back} uses Fast Point Feature Histograms (FPFH) \cite{rusu2009fast} descriptors and 2D pretrained models to extract multimodal features and construct a memory bank for anomaly detection. Further related work will be detailed in Section \ref{related}.

\begin{figure}[t]
    \centerline{\includegraphics[width = \linewidth]{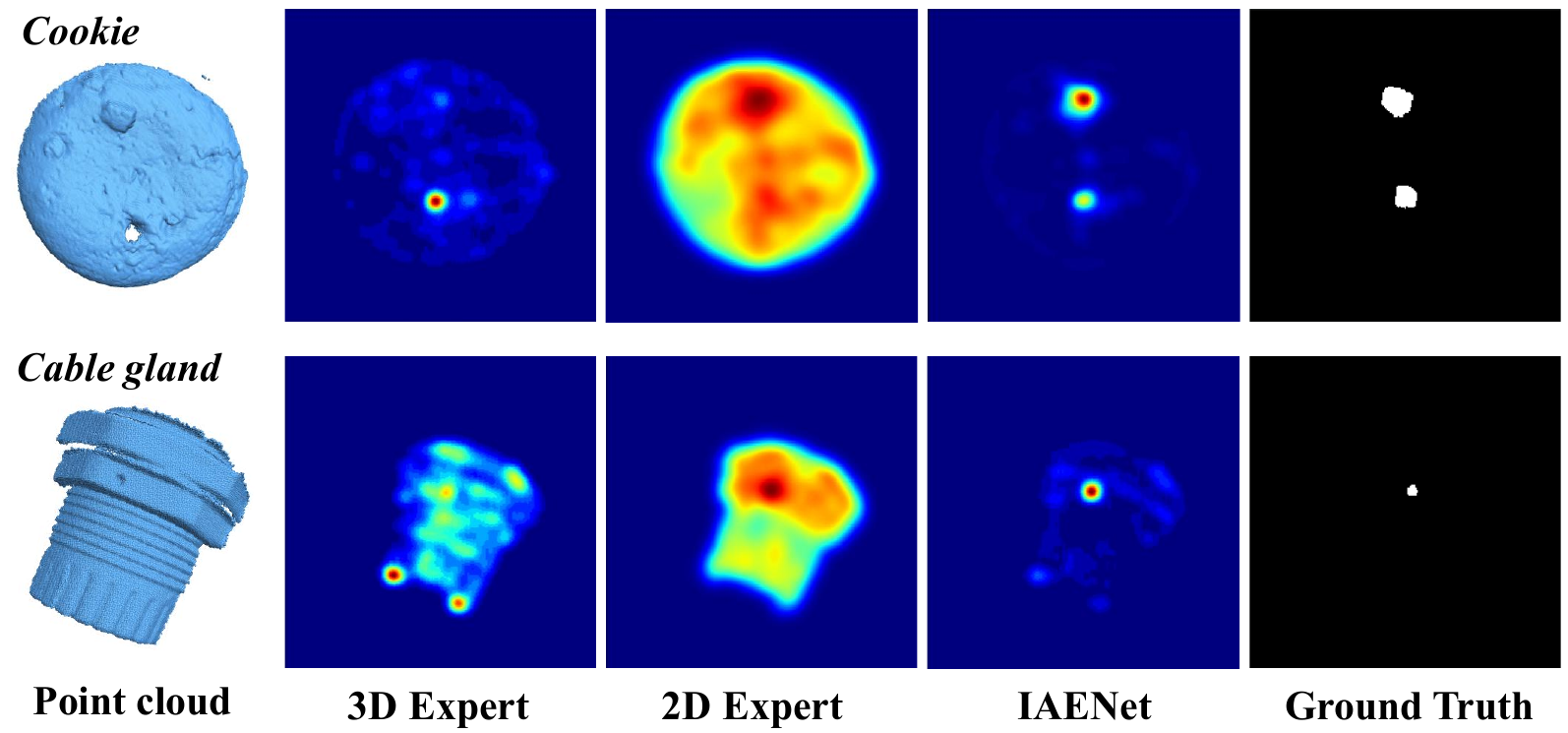}}
    \caption{The illustration of the anomaly score maps of different expert models on two representative objects from MVTec 3D-AD \cite{bergmann2021mvtec}. Despite receiving the same point cloud input, the 2D and 3D experts exhibit complementary strengths. When an anomaly is detected by one expert but not by another, our IAENet integrates both insights to accurately identify all anomalies. It also effectively suppresses the anomaly scores of normal points, thereby reducing the false positive rate.}
    \label{motivation}
\end{figure}

Given the current focus on multimodal data fusion in research, we have observed that anomaly detection based solely on point cloud data is underexplored. However, in many scenarios, point cloud data does not have corresponding RGB images available. Unfortunately, current approaches do not fully exploit the abundant geometric information contained within 3D point clouds. The primary problem is the lack of pretrained foundation models in the 3D domain, akin to those in the 2D domain, which possess powerful feature representation capabilities. This gap motivates us to adapt 2D pretrained foundation models by projecting 3D point clouds into textured 2D depth images, and integrate them with 3D expert models to construct an ensemble framework, termed as Importance-Aware Ensemble Network (IAENet), thereby enhancing 3D anomaly detection. As shown in Fig. \ref{motivation}, when fed with textured depth maps, the 2D expert is capable of effectively extracting rich semantic information from fine details, tending to identify all potential anomalies, which results in higher scores for normal points. However, the 3D expert captures more global information and struggles to detect subtle anomalies such as tiny holes in cable gland. Our IAENet model, on the other hand, effectively integrates the predictions from both experts, yielding more accurate detection results.

However, effectively integrating anomaly detection predictions from different source models remains a challenge. Although numerous multimodal studies have explored feature-level fusion, decision-level fusion, which involves the combination of predictions from different modalities, is often simplistic through basic strategies such as addition \cite{horwitz2023back} or taking the maximum value \cite{chu2023shape} for anomaly detection. These straightforward fusion tactics can be significantly undermined when there is a substantial discrepancy between the predictions of different modalities. This can lead to the fusion decision being adversely affected by the poorer-performing modality, resulting in a combined performance that is inferior to that of the individual modalities operating independently.

The main limitation of existing methods is that they assume equal contributions from the predictions of different models to the final results of anomaly detection. In reality, different models exhibit varying strengths across different scenarios (as demonstrated in Fig. \ref{motivation}), meaning that existing methods cannot adaptively determine the importance under different conditions. To address this, inspired by the research on feature selection \cite{yoon2018invase} which focuses on identifying important features, this paper proposes a novel Importance-Aware Fusion (IAF) module. The IAF module evaluates the contributions of anomaly scores from different models to the final anomaly detection, leading to more effective and robust anomaly detection results.

In general, we propose IAENet, a unified ensemble model that incorporates an innovative decision-level fusion module known as IAF for 3D point cloud-based anomaly detection. Specifically, the model begins with a phase of expert learning, where it trains a set of complementary 2D and 3D expert models to produce per-point predictions. These predictions are derived through feature extraction and a dual memory-bank retrieval process. A lightweight selector network then evaluates the contributions of these predictions to the anomaly detection task and assigns importance scores accordingly. These scores are used to weight the original predictions from the source models, which are then fed into a predictor network for nonlinear mapping to produce the final decision. Our IAENet not only integrates the collective knowledge of the source models but also preserves their unique strengths, thereby enhancing the overall performance of 3D anomaly detection.

To sum up, the contributions of this paper are as follows:
\begin{itemize}
    \item We propose IAENet, a novel ensemble framework that combines a 2D pretrained foundation model with a dedicated 3D expert to improve both accuracy and robustness in point cloud-based anomaly detection.
    
    \item We introduce an innovative Importance-Aware Fusion (IAF) module capable of adaptively evaluating the contributions of predictions from different models to the anomaly detection task. This module aims to preserve the unique strengths of individual models while combining the collective knowledge of the source experts.
    
    \item We design critical loss functions to ensure that the IAF module can accurately estimate the contributions of different models and obtain correct predictive results. This is essential for the IAF module to learn the appropriate importance scores and effectively integrate the information from various sources.

    \item Our IAENet achieves state-of-the-art results on the MVTec 3D-AD dataset, demonstrating a lower false positive rate, which is particularly valuable in industrial applications.
\end{itemize}

This paper is organized as follows: Section \ref{related} provides a comprehensive review of the related literature, Section \ref{method} details the methodology of our IAENet, Section \ref{exps} presents the experimental setup and results, and Section \ref{conclusion} concludes the paper with a summary of our findings and contributions.

\section{Related work}
\label{related}

In the related work section, we provide a comprehensive review that covers 2D anomaly detection, 3D anomaly detection, and score map fusion strategies.

\subsection{2D Anomaly Detection}
Anomaly detection based on 2D images has been extensively researched \cite{liu2025memory,li2025multi}, with the majority of studies validating their approaches on the MVTec-AD dataset \cite{bergmann2019mvtec}. In general, these methods can be broadly categorized into three main types: normalizing flow-based, knowledge distillation-based and memory bank-based methods. We will briefly introduce them as follows.

Normalizing flow-based methods \cite{gudovskiy2022cflow,bae2023pni} utilize normalizing flows to model the distribution of normal data. Anomalies are detected as data points that lie far from the learned distribution. The ability to model complex distributions makes these methods effective for capturing various anomalies.
Knowledge distillation techniques \cite{bergmann2020uninformed,rudolph2023asymmetric,zhang2023destseg} aim to transfer knowledge from a pretrained teacher model to a student model, which is then fine-tuned on normal data. The predictions of the student model are used to identify anomalies by highlighting discrepancies between its outputs and those of the teacher model.
Memory bank-based approaches \cite{roth2022towards} involve storing normal features extracted from training data into a memory bank. During inference, the features of test samples are compared against this bank to identify anomalies. This category of methods relies on the assumption that anomalies will differ significantly from the normal patterns stored in the memory bank. Due to robustness and adaptability \cite{tao2025g}, these approaches have become a foundation for the majority of current 3D anomaly detection methods \cite{horwitz2023back,chu2023shape,wang2023multimodal}. In addition to models trained from scratch, recent studies processing RGB images have leveraged pretrained foundation models such as ResNet \cite{he2016deep} and vision transformers \cite{dosovitskiy2020vit}, which have demonstrated effectiveness in anomaly detection \cite{cao2022informative,ammar_foundation_2025}.

While numerous 2D anomaly detection methods have been developed, they cannot be directly applied to 3D detection problems due to the inherent differences in data representation and complexity \cite{du20253d}.

\subsection{3D Anomaly Detection}

Unsupervised anomaly detection based on 3D point clouds has been under-researched until the emergence of the MVTec 3D-AD dataset \cite{bergmann2021mvtec}. However, since the MVTec 3D-AD dataset provides RGB images, most methods have focused on multimodal fusion research. Our discussion here will be limited to their 3D point cloud-based branches.

3D-ST \cite{bergmann2023anomaly} employs a teacher-student network architecture, utilizing RandLA-Net \cite{hu2020randla} as the point encoder to handle high-density data. BTF \cite{horwitz2023back} leverages local descriptors like FPFH to extract geometric information from 3D point clouds, also modeling the distribution of normal features through a memory bank, achieving satisfactory results. CPMF \cite{cao2024complementary} projects 3D point clouds and renders them into multiple depth maps, using 2D pretrained models to extract features, and then constructs a memory bank for anomaly detection. M3DM \cite{wang2023multimodal} introduces a pretrained PointMAE \cite{pang2022masked} as the 3D encoder for anomaly detection. Shape-Guided \cite{chu2023shape} uses neural implicit functions of signed distance fields \cite{ma2020neural} to represent local shapes of 3D point cloud and defines a new anomaly metric via signed distance function (SDF) functions, attaining the current state-of-the-art results in 3D anomaly detection.

\subsection{Score Map Fusion}

In the realm of multimodal methods, the strategies employed for score map fusion are relatively simplistic. The Shape-Guided \cite{chu2023shape} utilizes a maximum strategy, in which the final fused score map is composed by taking the maximum pixel-wise values from the score maps derived from the SDF and RGB modalities. 
BTF \cite{horwitz2023back} concatenates features from RGB and point cloud modalities and then performs detection based on a memory bank, which is equivalent to a linear score fusion \cite{tao2025g}.
M3DM \cite{wang2023multimodal}, on the other hand, adopts a data-driven method for fusion, employing a one-class Support Vector Machine (OCSVM) to achieve this integration. However, these methods fail to effectively evaluate the contributions of different modalities to the final anomaly detection outcome. Consequently, the fusion results can be significantly impacted by the poorer-performing modality.

Despite the introduction of a 3D pretrained model PointMAE in M3DM \cite{wang2023multimodal}, its performance in point-level anomaly localization does not surpass that of traditional FPFH descriptors. The primary reason may be that the scale of the ShapeNet dataset \cite{chang2015shapenet}, used for pretraining, is significantly smaller than ImageNet \cite{deng2009imagenet}, leading to a less powerful feature representation capability in the pretrained model. Therefore, the objective of this paper is to develop a novel ensemble model that integrates the collective knowledge of 2D pretrained foundation models and the currently best-performing 3D anomaly detection expert \cite{chu2023shape}, PointNet \cite{qi2017pointnet} combined with SDF. Additionally, a novel decision fusion module is designed to preserve the unique strengths of different models. This strategy aims to harness the complementary advantages of both 2D and 3D models, thereby enhancing the overall performance of anomaly detection.

\section{Methodology}
\label{method}

\subsection{Overview}
\label{overview}

\begin{figure*}[t]
    \centerline{\includegraphics[width = 0.9\linewidth]{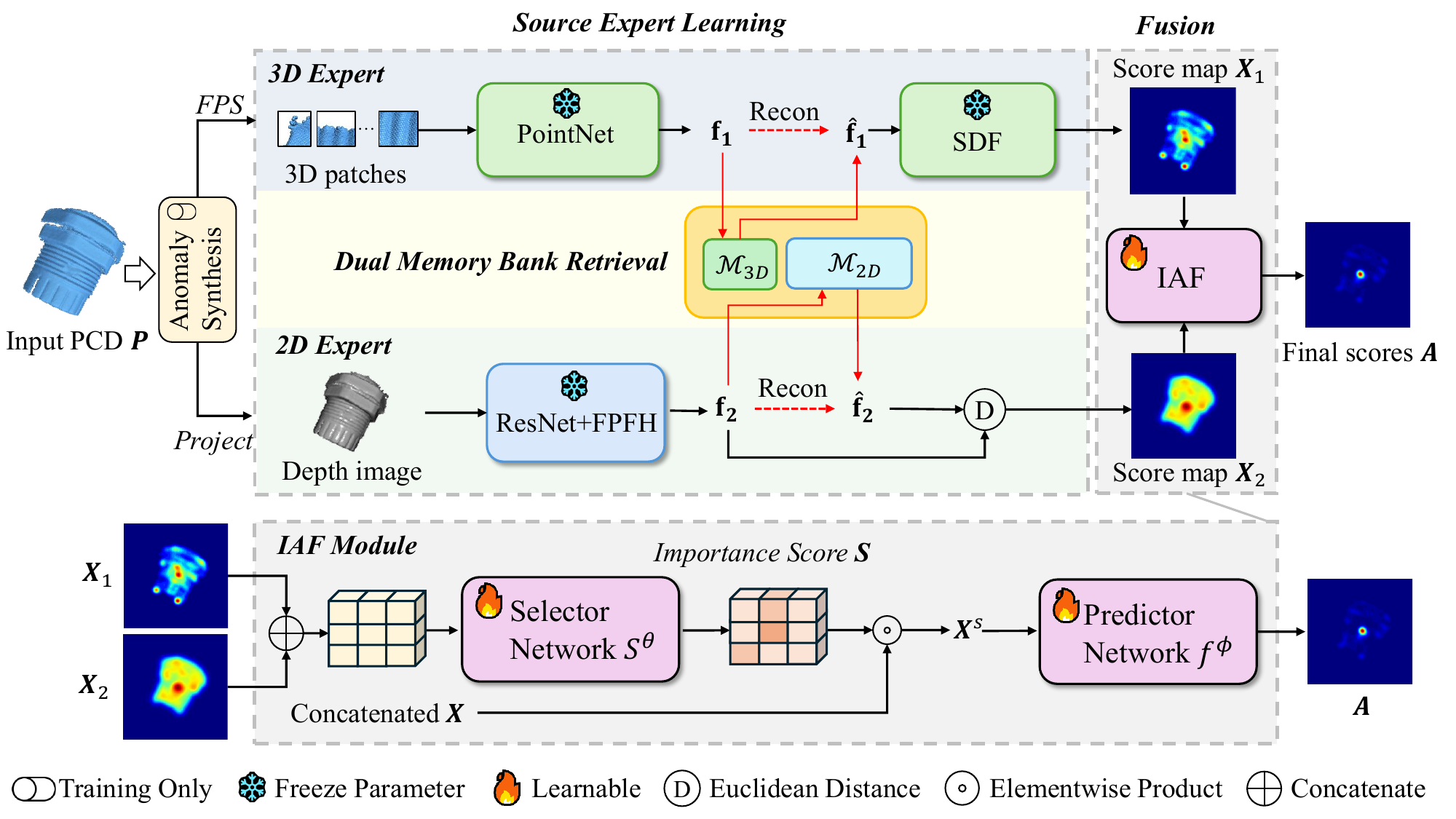}}
    \caption{The framework of our IAENet. During training, normal samples first pass through an anomaly synthesis module that generates pseudo anomalies (disabled at inference). The resulting anomalous point clouds are then processed by two source experts, yielding anomaly score maps $\mathbf{X_1}$ and $\mathbf{X_2}$, respectively. Finally, the Importance-Aware Fusion (IAF) module adaptively reweights these scores and outputs the refined anomaly map $\mathbf{A}$, effectively capitalizing on the complementary merits of both experts while suppressing their individual weaknesses.}
    \label{framework}
\end{figure*}

Given a point cloud $\mathbf{P} \in \mathbb{R}^{M \times 3}$ representing an arbitrary geometric surface, where $M$ denotes the number of points, the objective of point-level anomaly detection or localization is to localize the anomaly positions, i.e., to output anomaly scores $\mathbf{A} \in \mathbb{R}^M$ for all points. Here, anomalous points are assigned higher scores compared to normal points. Additionally, object-level anomaly detection aims to output a single anomaly score $s \in \mathbb{R}$ for the entire point cloud. In this paper, we focus primarily on the unsupervised anomaly detection task, which is more aligned with practical manufacturing scenarios. Specifically, we have access to only an anomaly-free point cloud dataset $\mathcal{D}=\{\mathbf{P}\}$ for training. During the testing phase, the model is expected to directly predict the point-level anomaly scores $\mathbf{A}$ and the object-level anomaly score $s$ for testing point cloud data.

As illustrated in Fig.~\ref{framework}, our IAENet comprises two steps: (i)~\emph{source expert learning} and (ii)~\emph{importance-aware fusion~(IAF)}. The first step trains complementary 2D and 3D experts on $\mathcal{D}$, producing per-point predictions via feature extraction and dual memory-bank retrieval (detailed in Sec.~\ref{expert learning}). The second step feeds these predictions into a \emph{selector network} that quantifies each expert’s reliability for every point, producing importance scores. A \emph{predictor network} then leverages these weights to fuse the expert outputs, preserving their unique strengths while suppressing weaknesses.

To optimize IAF, we synthesize an augmented dataset $\mathcal{D}' = \{(\mathbf{P}'_i, \mathbf{Y}_i)\}_{i=1}^{N}$ by injecting controlled anomalies into $\mathcal{D}$ using a Cut-Paste strategy~\cite{tao2025g}. Here, \(\mathbf{Y}_i\) represents the labels corresponding to \(\mathbf{P}'_i\), ranging from 0 to \(c\), where \(c=1\), indicating the presence or absence of anomalies. This supervision allows to learn robust integration without ever seeing real anomalies during training.

\subsection{Source Expert Learning}
\label{expert learning}

\subsubsection{3D Expert $\mathcal{E}_{3d}$}
\label{3D expert}

For the 3D expert, we adopt a PointNet–SDF pipeline~\cite{chu2023shape}.  
PointNet \cite{qi2017pointnet} acts as a patch-wise feature extractor that is first pretrained on the anomaly-free set~$\mathcal{D}$ by implicit surface reconstruction.  
Each patch of the input cloud~$\mathbf{P}$ is mapped to a latent vector $\mathbf{f}_1\in\mathbb{R}^{d_1}$.  
Given a query point~$\mathbf{q}$, the latent code and the query are concatenated and fed into a lightweight decoder~$\psi$ that predicts the signed distance

\begin{equation}
    s = \psi(\mathbf{q}, \mathbf{f}_1).
    \label{eq:sdf}
\end{equation}

During pretraining, the decoder is optimized to drive $s$ to zero on the manifold, thus learning an accurate normal surface representation.  
At anomaly detection step, the 3D expert operates with frozen parameters, and the magnitude $|s|$ serves as the anomaly score for each point, with larger values indicating stronger deviation from the learned normal geometry.

\subsubsection{2D Expert $\mathcal{E}_{2d}$}
\label{2D expert}
To effectively leverage the powerful feature representation capabilities of 2D foundation models, we adopt a frozen ResNet encoder as our 2D expert. Each point cloud is first rendered into a depth map via normal-based depth rendering~\cite{cao2024complementary}, which faithfully preserves fine-grained geometric textures, offering richer information than simple depth projection methods. This textured depth map is then processed by the pretrained ResNet to extract global semantic features. Finally, the global representation is concatenated with per-point FPFH descriptors to yield the composite feature $\mathbf{f}_2 \in \mathbb{R}^{d_2}$ that encodes both global context and local geometry.

\subsubsection{Dual Memory Bank Retrieval}
\label{sec:mem bank}

\begin{figure}[t]
    \centerline{\includegraphics[width = \linewidth]{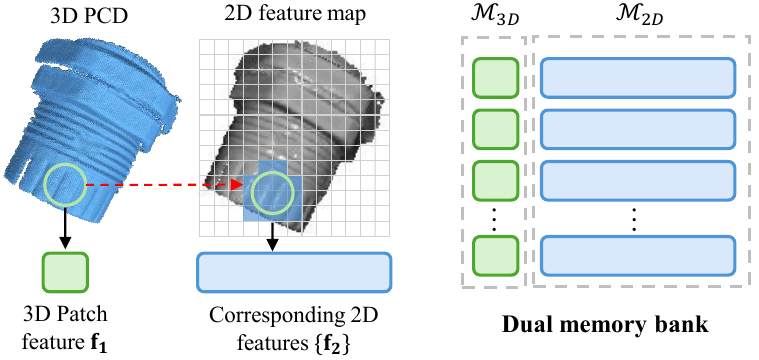}}
    \caption{Illustration of the construction of shape-guided dual memory bank.}
    \label{fig:mem bank}
\end{figure}

Following PatchCore~\cite{roth2022towards}, IAENet first populates two dedicated memory banks $(\mathcal{M}_{\text{3D}}, \mathcal{M}_{\text{2D}})$ with features extracted from the anomaly-free set $\mathcal{D}$.  
Instead of PatchCore's coreset subsampling, we adopt shape-guided retention~\cite{chu2023shape} to preserve semantically salient features.

\textbf{Population.}  
For every 3D patch feature $\mathbf{f}_1$ retained in $\mathcal{M}_{\text{3D}}$, we store the corresponding 2D features $\{\mathbf{f}_2\}$ of all points within its receptive field into $\mathcal{M}_{\text{2D}}$, as illustrated in Fig. \ref{fig:mem bank}.  
This cross-modal pairing guarantees that fine-grained geometry ($\mathbf{f}_1$) and global contextual cues ($\mathbf{f}_2$) are jointly indexed.

\textbf{Inference.}  
Given a test sample, we process it through both the 3D and 2D branches as follows:
\begin{itemize}
\item \textit{3D branch:} patch feature $\mathbf{f}_1$ retrieves $k_1$ nearest neighbors from $\mathcal{M}_{\text{3D}}$ to yield a reconstructed $\hat{\mathbf{f}}_1$.  
A query point $\mathbf{q}$ in the patch neighborhood is then processed by Eq.~\eqref{eq:sdf}, producing anomaly map $\mathbf{X}_1$.
\begin{equation}
\mathbf{X}_1(\mathbf{q}) = \bigl|\psi\bigl(\mathbf{q},\hat{\mathbf{f}}_1\bigr)\bigr|.
\label{eq:x1}
\end{equation}

\item \textit{2D branch:} for each point $\mathbf{p}$, its 2D feature $\mathbf{f}_2(\mathbf{p})$ searches $k_2$ nearest neighbors in $\mathcal{M}_{\text{2D}}$ to obtain the reconstructed feature $\hat{\mathbf{f}}_2(\mathbf{p})$.  
The anomaly score is
\begin{equation}
\mathbf{X}_2(\mathbf{p}) = \lVert\mathbf{f}_2(\mathbf{p}) - \hat{\mathbf{f}}_2(\mathbf{p})\rVert_2.
\label{eq:x2}
\end{equation}
\end{itemize}

This dual retrieval strategy exploits complementary 2D and 3D cues while keeping the banks compact and interpretable.

For the object-level anomaly score, the maximum values of the score maps are taken, i.e.,
\begin{equation}
    s_1 = \max(\mathbf{X}_1), \quad s_2 = \max(\mathbf{X}_2).
\end{equation}
Here, $s_1$ and $s_2$ represent the object-level anomaly scores derived from the score maps $\mathbf{X}_1$ and $\mathbf{X}_2$, respectively.

\subsection{Importance-Aware Fusion (IAF)}

\begin{figure}[t]
    \centerline{\includegraphics[width = \linewidth]{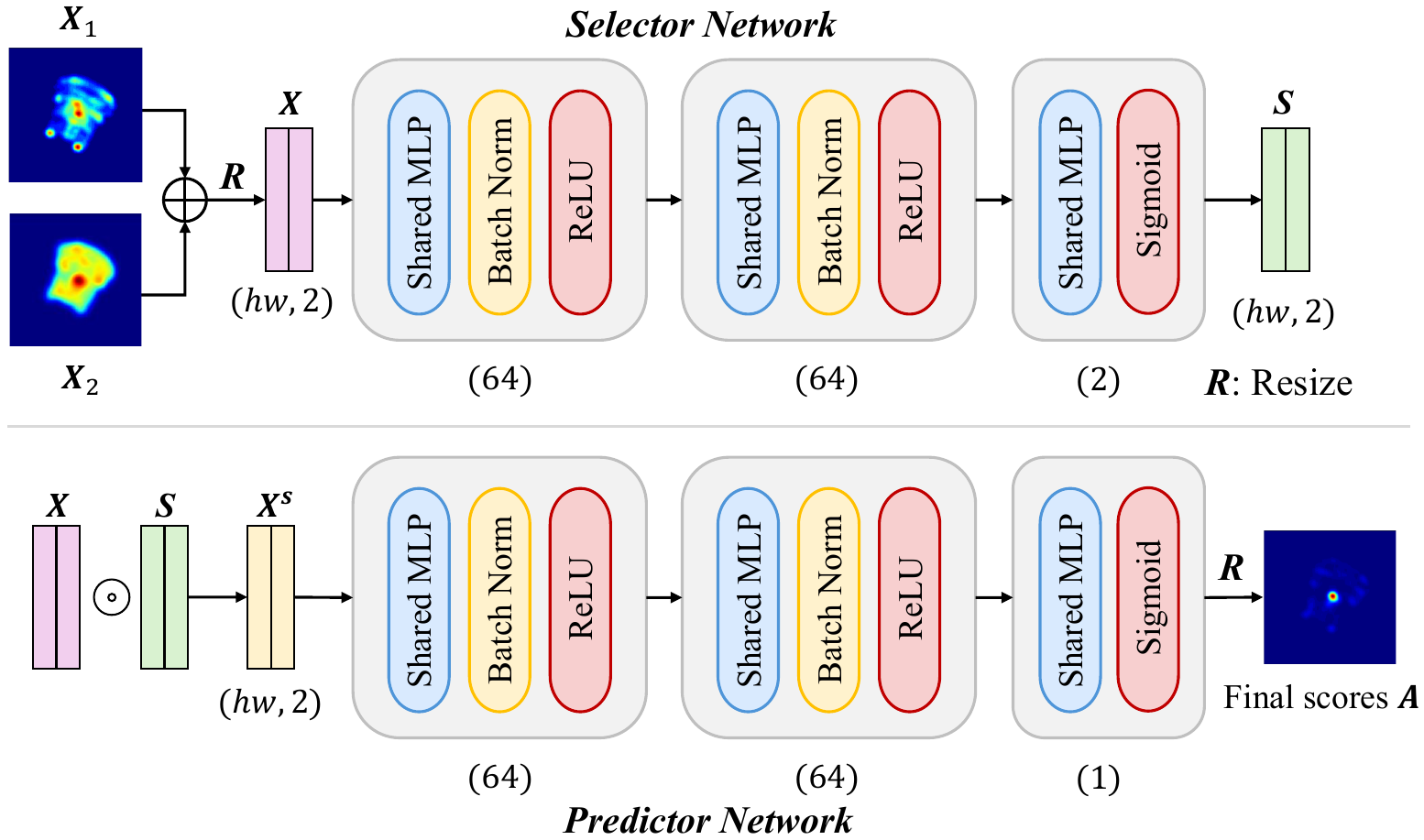}}
    \caption{The architecture of proposed selector network and predictor network.}
    \label{network}
\end{figure}

Given the prior anomaly maps $\mathbf{X}_1 \in \mathbb{R}^{h \times w}$ and $\mathbf{X}_2 \in \mathbb{R}^{h \times w}$ obtained from source expert learning, the objective is to effectively combine these maps to achieve superior detection results. However, existing methods often fail to accurately assess the reliability of anomaly detection results from different experts. When there is a significant disparity between the two maps, the final result can be severely skewed by the poorer-performing map. 

This limitation motivates the development of our IAF module $\mathcal{F}$. IAF learns an adaptive weighting that quantifies the trustworthiness of every spatial location in $\mathbf{X}_1$ and $\mathbf{X}_2$. These learned importance scores are then used to reweight and merge the two maps, allowing the ensemble to leverage their complementary strengths while protecting the output from deficiencies of any individual expert.

\subsubsection{Architecture of $\mathcal{F}$}

The IAF model $\mathcal{F}$ consists of two main components: the selector network $S^\theta$ and the predictor network $f^\phi$. The selector network $S^\theta$ is constructed using a set of shared multilayer perceptrons (MLPs) \cite{qi2017pointnet}, with the specific architecture depicted in Fig. \ref{network}. Its primary function is to evaluate the contributions of the two input score maps $\mathbf{X}_{1}$ and $\mathbf{X}_{2}$ (both resized into vectors $\mathbf{X}_{1}, \mathbf{X}_{2} \in \mathbb{R}^{hw \times 1}$) to the final anomaly detection task and to produce corresponding importance scores $\mathbf{S} \in \mathbb{R}^{hw\times2}$ that act as weights. This process is formulated as
\begin{equation}
    \mathbf{S}=S^\theta(\mathbf{X}_{1} \oplus \mathbf{X}_{2})=S^\theta(\mathbf{X}),
\end{equation}
where $\oplus$ is the concatenation.

The predictor network $f^\phi$ then takes the weighted score map as input and performs a nonlinear transformation to obtain the final fused anomaly score. Specifically, the weighted score map $\mathbf{X^S}$ is computed by multiplying each input score map $\mathbf{X}$ with its corresponding importance score $\mathbf{S}$ generated by the selector network. This weighted score map $\mathbf{X^S}$ is then fed into the predictor network, which consists of multiple nonlinear layers designed to capture complex patterns and interactions within the data, as shown in Fig. \ref{network}. The output of the predictor network is the final anomaly score $\mathbf{A}$.
\begin{equation}
    \mathbf{A} =f^\phi(\mathbf{X}, \mathbf{S})=f^\phi(\mathbf{X} \odot \mathbf{S})=f^\phi(\mathbf{X^\mathbf{S}}),
\end{equation}
where $\odot$ is elementwise product.

\subsubsection{Loss Functions}

\textbf{Problem formulation}. The objective of this study is to develop an importance-aware fusion module $\mathcal{F}$ that outperforms individual experts $\mathcal{E}_{2d}$ and $\mathcal{E}_{3d}$ in anomaly detection task. This is formalized as
\begin{equation}
    \mathcal{P}(\mathcal{F},\mathbf{Y}) > \mathcal{P}(\mathcal{E}_{2d},\mathbf{Y}), \mathcal{P}(\mathcal{E}_{3d},\mathbf{Y}),
\end{equation}
where $\mathcal{P}$ denotes the performance of the model relative to target labels $\mathbf{Y}$. Performance is quantified using cross-entropy ($\mathrm{CE}$) \cite{wan2024knowledge}, leading to the constraint
\begin{equation}
    \mathrm{CE}(\mathcal{F},\mathbf{Y}) < \mathrm{CE}(\mathcal{E}_{2d},\mathbf{Y}), \mathrm{CE}(\mathcal{E}_{3d},\mathbf{Y}),
\end{equation}
or equivalently,
\begin{equation}
    \mathcal{C}_{\mathcal{F}} < b = \min(\mathcal{C}_{2d}, \mathcal{C}_{3d}).
\label{eq:obj}
\end{equation}
Here, $\mathcal{C}_{\mathcal{F}}$ is the cross-entropy loss of the fused model, $\mathcal{C}_{2d}$ and $\mathcal{C}_{3d}$ are the baseline cross-entropy losses, and $b$ is the minimum of these two baselines. Note that $b$ is a constant, as the input score maps do not change during training.

\textbf{Selector loss.}
We reformulate the objective function \eqref{eq:obj} as an optimizable loss function:
\begin{equation}
    \mathcal{R} = \max(m - (b - \mathcal{C}_{\mathcal{F}}), 0),
\end{equation}
where $m$ denotes the margin parameter, and $\mathcal{C}_{\mathcal{F}}$ is defined as
\begin{equation}
    \mathcal{C}_{\mathcal{F}} = -\mathbb{E}_{\mathbf{X} \sim P} \left[ \sum_{i=1}^{c} y_i \log(p(y_i | \mathbf{X^S})) \right],
\end{equation}
where $\mathbf{X}$ is the random variable with density $P$, and the probability $p(y_i|\mathbf{X^S})$ is given by the predictor network $f^\phi$ as follows
\begin{equation}
    p(y_i|\mathbf{X^S}) = f^\phi_i(\mathbf{X},\mathbf{S}).
\end{equation}
Then, the stochastic form of $\mathcal{R}$ is
\begin{equation}
    r(\mathbf{X}, \mathbf{S})=\max(m-(b+\sum_{i=1}^cy_i\log(f^\phi_i(\mathbf{X},\mathbf{S}))),0).
\label{eq:re_stoch}
\end{equation}
So the final $\mathcal{R}$ is 
\begin{align}
    \mathcal{R}(S)&=\mathbb{E}_{\mathbf{X} \sim P}[r(\mathbf{X},S^\theta(\mathbf{X}))].
\label{eq:reward}
\end{align}

Inspired by reward-regularized feature selection in \cite{yoon2018invase}, we incorporate an entropy regularizer into the selector loss to encourage early exploration and later specialization.
The entropy over the selector’s soft-assignment $\mathbf{S}$ is
\begin{equation}
    \mathcal{H} = -\mathbf{S} \log(\mathbf{S}).
\label{eq:entropy}
\end{equation}
Combining this with $r(\cdot)$ from Eq.~\eqref{eq:reward} yields the loss
\begin{align}
    \mathcal{L}_{s}&=\mathbb{E}_{\mathbf{X} \sim P}[r(\mathbf{X},S^\theta(\mathbf{X})) \cdot \mathcal{H}] \\
    &=\sum_{\mathbf{x}} \frac{1}{N}r(\mathbf{x},S^\theta(\mathbf{x})) \cdot (-\mathbf{S} \log(\mathbf{S})).
\end{align}
Substituting Eq. \eqref{eq:re_stoch} gives 
\begin{equation}
\begin{split}
    \mathcal{L}_{s}&=-\frac{1}{N} \sum_\mathbf{x} \max \left( m - \left( b + \right.\right. \\&
    \left.\left. \sum_{i=1}^c y_i \log(f^\phi_i(\mathbf{x}, S^\theta(\mathbf{x}))) \right),  
    0 \right) \cdot \left(S^{\theta}(\mathbf{x}) \log(S^{\theta}(\mathbf{x}))\right)
\end{split}
\label{selector loss}
\end{equation}

The selector loss is designed to encourage early exploration and later exploitation, ensuring that the selector network can effectively evaluate and specialize in the most informative expert. Specifically, the optimization process of \( \mathcal{L}_{s} \) ensures that the weight combination \( \mathbf{S} \) starts from equal contributions, i.e., [0.5, 0.5]. As \( \mathcal{H} \) decreases, \( \mathbf{S} \) begins to approach [0, 1] or [1, 0]. This process guarantees exploration of all weight combinations. At this point, \( \mathcal{R} \) acts as a performance gate, determining when to cease exploration. Specifically, when \( \mathcal{H} \) drives the selector network to explore a particular combination, such as [0.2, 0.8], and \( \mathcal{R} = 0 \), it indicates that the current weight combination \( \mathbf{S} \) meets the requirements, and the gate closes, halting exploration and shifting to exploitation ($\mathcal{L}_{s} = 0$). 

In the context of anomaly detection tasks, when there is a significant discrepancy between the predictions of the two source experts, for instance, if the 3D expert's detection results are inaccurate, to prevent its poor predictions from affecting the final decision, \( \mathcal{H} \) will drive the selector network to output a weight combination \(\mathbf{S}\) closer to \( [0, 1] \). At this time, the performance gate will also close (with \( \mathcal{R} = 0 \)) later, potentially resulting in a weight combination approaching [0, 1], such as [0.1, 0.9], thus assigning a smaller weight to the 3D expert, ensuring that the final prediction is not adversely influenced by errors.

\textbf{Predictor loss.} To ensure that the predictor network can perform the correct nonlinear mapping based on the importance score $\mathbf{S}$ output by the selector, we employ cross-entropy for separate training. This loss function is defined as
\begin{equation}
    \mathcal{L}_{p} = -\sum_{\mathbf{x}} \sum_{i=1}^c y_i\log(f^\phi_i(\mathbf{x} \odot S^\theta(\mathbf{x})))).
\end{equation}
By minimizing this loss, the predictor network learns to effectively combine the weighted score maps into a fused score map $\mathbf{A}$ that accurately reflects the likelihood of anomalies in the input data.

The overall loss function is a combination of the predictor loss $\mathcal{L}_{p}$ and the selector loss $\mathcal{L}_{s}$. Specifically, the overall loss $\mathcal{L}_{\text{final}}$ is defined as
\begin{equation}
    \mathcal{L}_{\text{final}} = \mathcal{L}_{p} + \lambda \mathcal{L}_{s},
\label{final loss}
\end{equation}
where $\lambda$ is a hyperparameter that balances the trade-off between the predictor loss and the selector loss. This combined loss function ensures that the model not only optimizes the accuracy of the anomaly detection results but also balances the contributions of different experts through the selector network.

Finally, the fusion of the object-level scores $s_1 \in \mathbb{R}$ and $s_2 \in \mathbb{R}$ follows a similar procedure. First, the scores $s_1$ and $s_2$ are concatenated and fed into the same selector network to obtain a weighted score. This weighted score is then passed through the predictor network for nonlinear transformation to produce the final score. Mathematically, this process is expressed as
\begin{equation}
    s = f^\phi(s_1 \oplus s_2, \, S^\theta(s_1 \oplus s_2)).
\end{equation}
The training and inference of our methodology is summarized in Algorithm \ref{alg}.

\begin{algorithm}
\caption{Model training and inference.}\label{alg}
\begin{algorithmic}[1]
\renewcommand{\algorithmicrequire}{\textbf{Stage 1:}}
\REQUIRE \textbf{Model training}
\renewcommand{\algorithmicrequire}{\textbf{Input:}}
\REQUIRE Normal data \( \mathcal{D} \), synthetic data \( \mathcal{D}^\prime \), margin \( m \), \( \lambda \)
\renewcommand{\algorithmicensure}{\textbf{Output:}}
\ENSURE Dual memory bank \( (\mathcal{M}_{2D}, \mathcal{M}_{3D}) \), IAF model \( \mathcal{F} \)
\State Initialize memory banks \( (\mathcal{M}_{2D}, \mathcal{M}_{3D}) \)
\FOR{ \( \mathbf{P} \) in \( \mathcal{D} \) }
    \State $(\mathbf{f}_{1}, \mathbf{f}_{2}) \leftarrow$ Extract features from $\mathbf{P}$ using $\mathcal{E}_{3d}$ and $\mathcal{E}_{2d}$.
    \State Update dual memory bank \( (\mathcal{M}_{2D}, \mathcal{M}_{3D}) \) with features \( (\mathbf{f}_{1}, \mathbf{f}_{2}) \).
\ENDFOR
\REPEAT
    \FOR{ \( \mathbf{P}^\prime \) in \( \mathcal{D}^\prime \) }
        \State $(\mathbf{f}_{1}', \mathbf{f}_{2}') \leftarrow$ Extract features from $\mathbf{P}^\prime$ using $\mathcal{E}_{3d}$ and $\mathcal{E}_{2d}$.
        \State $(\mathbf{X}_1, \mathbf{X}_2, s_1, s_2) \leftarrow$ Calculate prior anomaly scores using $(\mathcal{M}_{2D}, \mathcal{M}_{3D})$.
        \State \( (\mathbf{X}, s) \leftarrow \mathcal{F}(\mathbf{X}_1, \mathbf{X}_2, s_1, s_2) \).
        \State Compute final loss Eq. \eqref{final loss} and update \( \mathcal{F} \).
    \ENDFOR
\UNTIL{convergence}
\renewcommand{\algorithmicrequire}{\textbf{Stage 2:}}
\REQUIRE \textbf{Model inference}
\renewcommand{\algorithmicrequire}{\textbf{Input:}}
\REQUIRE Testing point cloud \( \mathbf{P}_t \)
\renewcommand{\algorithmicensure}{\textbf{Output:}}
\ENSURE Fused anomaly scores \( (\mathbf{A}, s) \)
\State Load the parameters of IAF \( \mathcal{F} \)  and \( (\mathcal{M}_{2D}, \mathcal{M}_{3D}) \).
\State \( (\mathbf{f}_{1}^t, \mathbf{f}_{2}^t) \leftarrow \) Extract features from $\mathbf{P}_t$ using source experts.
\State $(\mathbf{X}^t_1, \mathbf{X}^t_2, s^t_1, s^t_2) \leftarrow$ Calculate prior anomaly scores using $(\mathcal{M}_{2D}, \mathcal{M}_{3D})$.
\State \( (\mathbf{A}, s) \leftarrow \mathcal{F}(\mathbf{X}^t_1, \mathbf{X}^t_2, s^t_1, s^t_2) \).
\end{algorithmic}
\end{algorithm}

\section{Experiments}
\label{exps}

\subsection{Experimental Settings}
\subsubsection{Datasets}
Our experiments are conducted on the MVTec 3D-AD \cite{bergmann2021mvtec} industrial dataset, which provides real scanned point cloud data for 10 types of objects. For each object category, there are 210 to 361 training samples, all of which are normal data, and 69 to 132 testing samples that include both normal and anomalous data. The anomalous samples encompass 4 to 5 types of anomalies. In summary, the dataset comprises 2,656 training samples and 1,197 testing samples.

In addition to real datasets, we also generated a synthetic dataset based on original MVTec 3D-AD to train the IAF module. Specifically, we first create an anomaly mask, then utilize a Cut-Paste technique \cite{tao2025g} to cut points corresponding to the mask from a source point cloud. The source point cloud could be derived from normal point clouds of other categories or general datasets. These cut points are then pasted onto the target point cloud and subjected to random transformations to produce the synthetic anomaly-enhanced dataset. For each class of objects, we generated 800 samples to serve as training data.

\subsubsection{Implementation Details}
\textbf{Data preprocessing}. The preprocessing steps for the source expert learning, such as background removal, point cloud patching, and depth map projection, are consistent with the methods described in Shape-Guided \cite{chu2023shape} and CPMF \cite{cao2024complementary}. Unlike the multi-view projection of CPMF, we only project a frontal view, thus avoiding the choice of projection angle affecting the detection results. These steps are crucial for preparing the data in a format suitable for effective feature extraction and anomaly detection. 

\textbf{Source expert learning.} For 3D expert, we adopt the identical architecture of PointNet and SDF in~\cite{chu2023shape}. PointNet outputs patchwise features $\mathbf{f}_1\in\mathbb{R}^{128}$.  Pretraining is performed on the anomaly-free set via SDF reconstruction loss.
For 2D expert, we employ a frozen Wide-ResNet-50-2 pretrained on ImageNet. Features from the first two layers are concatenated with FPFH descriptors, yielding an 801 dimensional representation $\mathbf{f}_2$.

\textbf{IAF training.} For the training of our Importance-Aware Fusion (IAF) model, we employ the AdamW optimizer, iterating over 150 epochs with a batch size of 32. The learning rate is set at 0.01, and we utilize a cosine annealing schedule to adjust the learning rate over the training process. In the loss function, we set the margin $m=0.1$ and the hyperparameter $\lambda=1$. It should be noted that these parameters may require fine-tuning for some categories with significantly varying inputs to balance the contribution of each source models and ensure effective convergence of the model. All experiments are conducted on a single RTX 4090 GPU.

\subsubsection{Evaluation Metrics}
\label{metrics}
For object-level anomaly detection, we employ the commonly used Area Under the Receiver Operating Characteristic curve (AUROC) metric, denoted as O-AUROC. For point-level anomaly detection, we utilize both the AUROC and the Area Under Per-Region Overlap curve (AUPRO) \cite{bergmann2021mvtec} for evaluation, denoted as P-AUROC and AUPRO, respectively. The AUPRO metric involves calculating the integration of the PRO values over the range of false-positive rates (FPRs). Similar to most previous methods, we set the upper limit of the FPR integration to the default value of 0.3, which is denoted as AUPRO@30\%. A smaller FPR integration limit implies a stricter tolerance for false positives.

\subsection{Comparison Results}

We compared our approach with other current state-of-the-art 3D anomaly detection methods, including Voxel GAN \cite{bergmann2021mvtec}, Voxel AE \cite{bergmann2021mvtec},
3D-ST \cite{bergmann2023anomaly}, BTF \cite{horwitz2023back}, AST \cite{rudolph2023asymmetric}, M3DM \cite{wang2023multimodal}, CPMF \cite{cao2024complementary}, and Shape-Guided \cite{chu2023shape}. Table \ref{O-AUROC} shows the performance of each method for object-level anomaly detection across all categories. Our method ranked second-best, with results very close to the best-performing CPMF (with a difference of just 0.006). However, the anomaly localization performance of CPMF is inferior to that of our approach, as will be illustrated in the subsequent sections. 
It is worth noting that CPMF relies on 27 viewpoint projections, whose angles and count must be carefully tuned—suboptimal choices can directly degrade performance. In contrast, our 2D expert achieves comparable object-level anomaly detection with just a single projection, avoiding these complications and demonstrating robustness and adaptability in diverse industrial scenarios. 

\begin{figure*}[t]
    \centerline{\includegraphics[width = 0.95\linewidth]{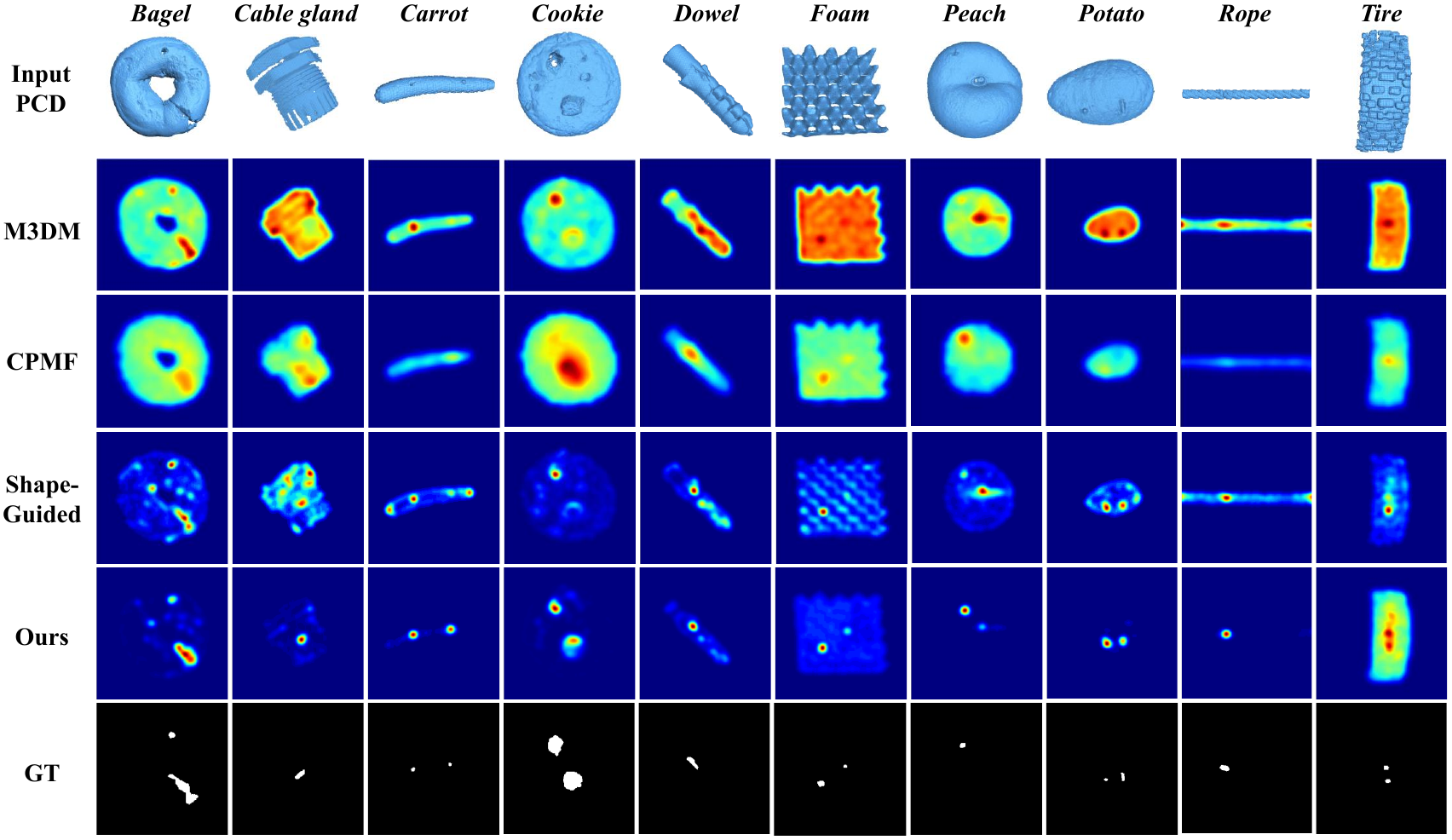}}
    \caption{Qualitative comparison of anomaly score maps on the MVTec 3D-AD dataset. This is the visualization of the final output of the anomaly score map for each methodology, where blue represents values close to 0 and red represents larger anomaly score values.}
    \label{comparison_vis}
\end{figure*}

\begin{table*}[]
\centering
\caption{\textbf{O-AUROC} scores for anomaly detection across various categories in the MVTec 3D-AD dataset (Best results in bold).}
\begin{tabular}{@{}lccccccccccc@{}}
\toprule
Method       & Bagel          & Cable gland    & Carrot         & Cookie         & Dowel          & Foam           & Peach          & Potato         & Rope           & Tire           & Mean           \\ \midrule
Voxel GAN    & 0.383          & 0.623          & 0.474          & 0.639          & 0.564          & 0.409          & 0.617          & 0.427          & 0.663          & 0.577          & 0.537          \\
Voxel AE     & 0.693          & 0.425          & 0.515          & 0.790          & 0.494          & 0.558          & 0.537          & 0.484          & 0.639          & 0.583          & 0.571          \\
BTF (SIFT)   & 0.711          & 0.656          & 0.892          & 0.754          & 0.828          & 0.686          & 0.622          & 0.754          & 0.767          & 0.598          & 0.727          \\
BTF (FPFH)   & 0.825          & 0.551          & 0.952          & 0.797          & 0.883          & 0.582          & 0.758          & 0.889          & 0.929          & 0.653          & 0.782          \\
AST          & 0.881          & 0.576          & 0.956          & 0.957          & 0.679          & 0.797          & 0.990          & 0.915          & 0.956          & 0.611          & 0.832          \\
M3DM         & 0.941          & 0.651          & 0.965          & 0.969          & 0.905          & 0.760          & 0.880          & \textbf{0.974} & 0.926          & 0.765          & 0.874          \\
CPMF         & 0.981          & 0.888          & \textbf{0.992} & 0.989          & 0.962          & \textbf{0.794} & 0.990          & 0.963          & \textbf{0.979} & 0.966          & \textbf{0.950} \\
Shape-Guided & \textbf{0.983} & 0.710          & 0.974          & \textbf{0.993} & \textbf{0.971} & 0.722          & \textbf{0.992} & 0.964          & 0.966          & 0.931          & 0.921          \\
Ours         & 0.969          & \textbf{0.954} & 0.990          & 0.945          & 0.955          & 0.756          & 0.983          & 0.966          & 0.954          & \textbf{0.968} & 0.944          \\ \bottomrule
\end{tabular}
\label{O-AUROC}
\end{table*}

Tables \ref{P-AUROC} and \ref{AUPRO} present the quantitative results for point-level anomaly localization, evaluated by P-AUROC and AUPRO, respectively. It is evident that our method achieves the best results (AUPRO = 0.944), significantly outperforming CPMF (AUPRO = 0.929), which is the top performer in the object-level anomaly detection task. Beyond the superior metrics, the visualization of the score maps in Fig. \ref{comparison_vis} demonstrates the strengths of our approach. First, the proposed IAENet not only accurately localizes anomalies but also effectively suppresses the anomaly scores in normal regions. This enhances the distinction between anomaly and normal regions, thereby reducing the false positive rate, which is of substantial significance in practical industrial scenarios.

\begin{figure}[]
    \centerline{\includegraphics[width = \linewidth]{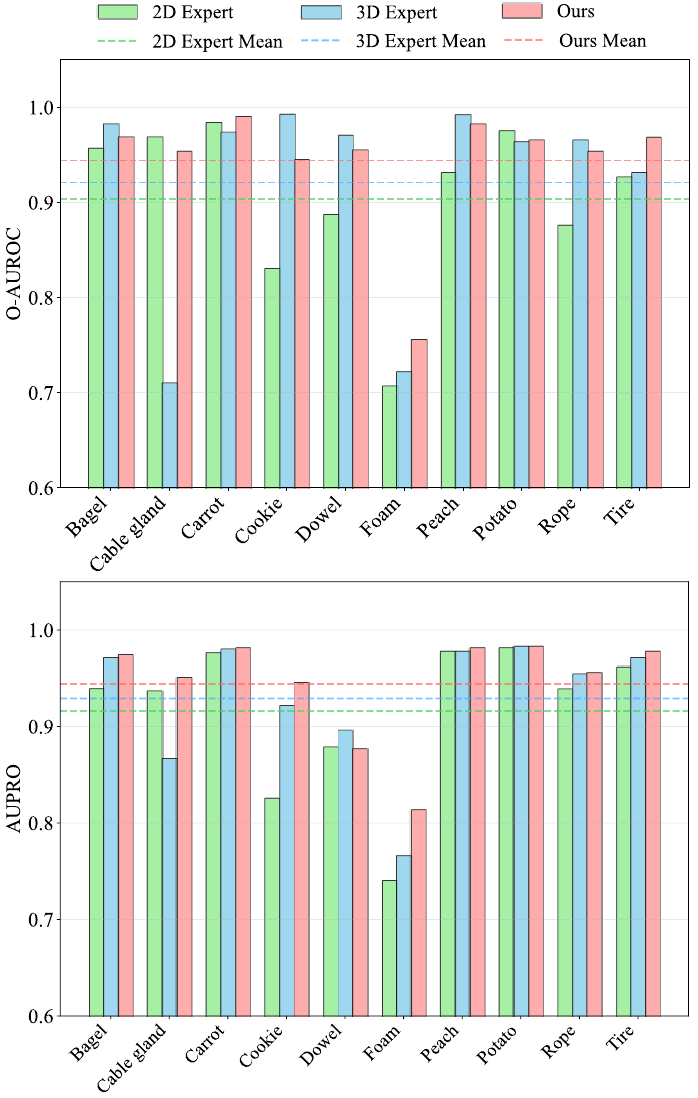}}
        \caption{Object-level (O-AUROC) and point-level (AUPRO) anomaly detection results on the MVTec 3D-AD dataset by the 2D expert, 3D expert, and our IAENet.}
    \label{hist}
\end{figure}

Fig. \ref{hist} displays the object-level anomaly detection and point-level anomaly localization results by the 2D expert, 3D expert, and our IAENet on the MVTec 3D-AD dataset. It is evident that across both object-level anomaly detection and point-level anomaly localization, IAENet (with mean O-AUROC of 0.944 and mean AUPRO of 0.944) outperforms individual expert models across all categories: the 2D expert achieves mean O-AUROC of 0.904 and mean AUPRO of 0.914, while the 3D expert attains mean O-AUROC of 0.921 and mean AUPRO of 0.929. Furthermore, it is clear that our IAF module effectively manages scenarios with significant discrepancies in the results of different sources of experts. For instance, in the object-level anomaly detection for the \textit{Cable gland} category, where the 2D expert scored 0.969 and the 3D expert scored 0.71, the IAF module still achieves a satisfactory result of 0.954, unaffected by the inferior performance. This confirms the efficacy of the proposed IAF module, demonstrating its ability to determine which source expert's results contribute more favorably to the final anomaly detection.

Another advantage of our IAENet is its capability to effectively detect subtle anomalies on complex surfaces. As depicted in Fig. \ref{comparison_vis}, our approach effectively identifies the scratch on the thread surface of a cable gland, an anomaly that other methods struggle to identify. The score map produced by our method distinctly marks the location of the anomaly, with anomaly scores for normal regions being close to zero. This demonstrates that our IAENet not only accurately locates subtle anomalies with high confidence but also accurately discerns complex normal surfaces.

\begin{table*}[]
\centering
\caption{\textbf{P-AUROC} scores for anomaly localization across various categories in the MVTec 3D-AD dataset (Best results in bold).}
\begin{tabular}{@{}lccccccccccc@{}}
\toprule
Method       & Bagel          & Cable Gland    & Carrot         & Cookie         & Dowel          & Foam           & Peach          & Potato         & Rope           & Tire           & Mean           \\ \midrule
BTF (SIFT)   & 0.974          & 0.862          & 0.993          & 0.952          & \textbf{0.980} & 0.862          & 0.955          & 0.996          & 0.993          & 0.971          & 0.954          \\
BTF (FPFH)   & \textbf{0.995} & 0.965          & \textbf{0.999} & 0.947          & 0.966          & 0.928          & 0.996          & \textbf{0.999} & \textbf{0.996} & 0.991          & 0.978          \\
M3DM         & 0.981          & 0.947          & 0.996          & 0.934          & 0.960          & \textbf{0.944} & 0.988          & 0.994          & 0.994          & 0.983          & 0.972          \\
CPMF         & 0.986          & 0.986          & 0.997          & 0.926          & 0.969          & 0.936          & 0.997          & 0.998          & \textbf{0.996} & 0.996          & 0.979          \\
Shape-Guided & 0.991          & 0.962          & 0.998          & 0.947          & 0.959          & 0.93           & 0.996          & \textbf{0.999} & 0.995          & 0.996          & 0.977          \\
Ours         & 0.994          & \textbf{0.988} & 0.998          & \textbf{0.964} & 0.944          & \textbf{0.944} & \textbf{0.998} & \textbf{0.999} & 0.994          & \textbf{0.997} & \textbf{0.982} \\ \bottomrule
\end{tabular}
\label{P-AUROC}
\end{table*}

\begin{table*}[]
\centering
\caption{\textbf{AUPRO@30\%} scores for anomaly localization across various categories in the MVTec 3D-AD dataset (Best results in bold).}
\begin{tabular}{@{}lccccccccccc@{}}
\toprule
Method       & Bagel          & Cable gland    & Carrot         & Cookie         & Dowel          & Foam           & Peach          & Potato         & Rope           & Tire           & Mean           \\ \midrule
Voxel GAN    & 0.440          & 0.453          & 0.875          & 0.755          & 0.782          & 0.378          & 0.392          & 0.639          & 0.775          & 0.389          & 0.583          \\
Voxel AE     & 0.260          & 0.341          & 0.581          & 0.351          & 0.502          & 0.234          & 0.351          & 0.658          & 0.015          & 0.185          & 0.348          \\
3D-ST        & 0.950          & 0.483          & \textbf{0.986} & 0.921          & 0.905          & 0.632          & 0.945          & \textbf{0.988} & \textbf{0.976} & 0.542          & 0.833          \\
BTF (SIFT)   & 0.942          & 0.842          & 0.974          & 0.896          & \textbf{0.910} & 0.723          & 0.944          & 0.981          & 0.953          & 0.929          & 0.909          \\
BTF (FPFH)   & 0.973          & 0.879          & 0.982          & 0.906          & 0.892          & 0.735          & 0.977          & 0.982          & 0.956          & 0.961          & 0.924          \\
M3DM         & 0.943          & 0.818          & 0.977          & 0.882          & 0.881          & 0.743          & 0.958          & 0.974          & 0.950          & 0.929          & 0.906          \\
CPMF         & 0.958          & 0.946          & 0.979          & 0.868          & 0.897          & 0.746          & 0.980          & 0.981          & 0.961          & 0.977          & 0.929          \\
Shape-Guided & 0.972          & 0.867          & 0.981          & 0.922          & 0.897          & 0.767          & 0.978          & 0.983          & 0.955          & 0.972          & 0.929          \\
Ours         & \textbf{0.975} & \textbf{0.951} & 0.982          & \textbf{0.946} & 0.877          & \textbf{0.814} & \textbf{0.982} & 0.983          & 0.956          & \textbf{0.978} & \textbf{0.944} \\ \bottomrule
\end{tabular}
\label{AUPRO}
\end{table*}

\subsection{Analysis}

\subsubsection{Difference Between Anomaly and Normal Points}
As demonstrated by the visualization results, one of the strengths of our IAENet is its ability to effectively suppress the anomaly scores of normal points, thereby magnifying the difference between anomalies and normal regions. In this section, we will quantify and analyze this advantage by examining the results of AUPRO at different integration limits and the distribution of anomaly scores.

\begin{figure}[t]
    \centerline{\includegraphics[width = 0.9\linewidth]{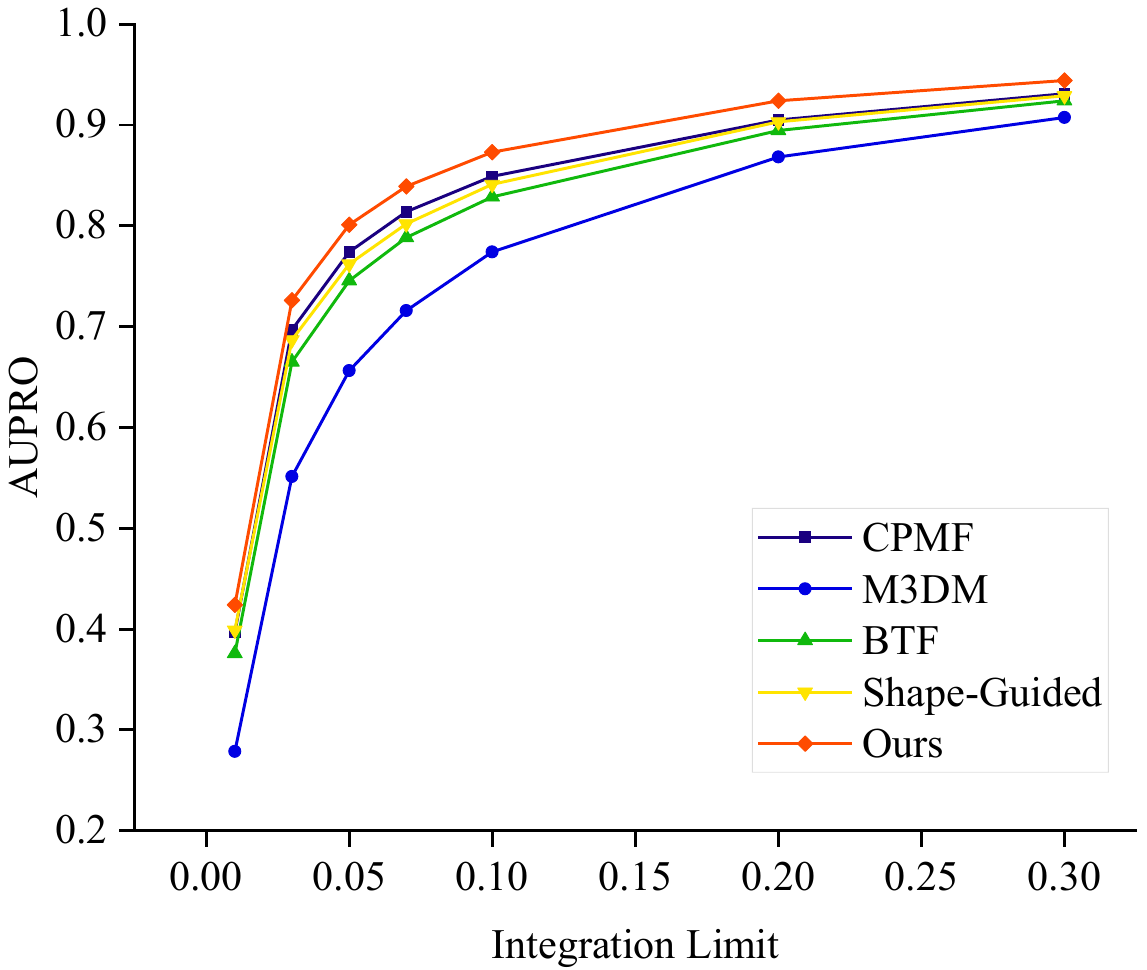}}
    \caption{Anomaly localization performance (AURPO) of our method with comparison methods at different integration limits.}
    \label{integration}
\end{figure}

\textbf{Comparison on different integration limits.} 
Our previous AUPRO metrics are conducted with an integration upper limit of 0.3, i.e., AUPRO@30\%, as mentioned in Section \ref{metrics}, where a smaller integration limit indicates a stricter tolerance for false positives. Therefore, we compared our IAENet with benchmarks across seven decreasing integration limits $\{0.3,0.2,0.1,0.07,0.05,0.03,0.01\}$ for the AUPRO results, as shown in Fig. \ref{integration}. It can be observed that across the various integration limits, our method consistently achieves the state-of-the-art performance.

\begin{figure}[t]
    \centerline{\includegraphics[width = 0.9\linewidth]{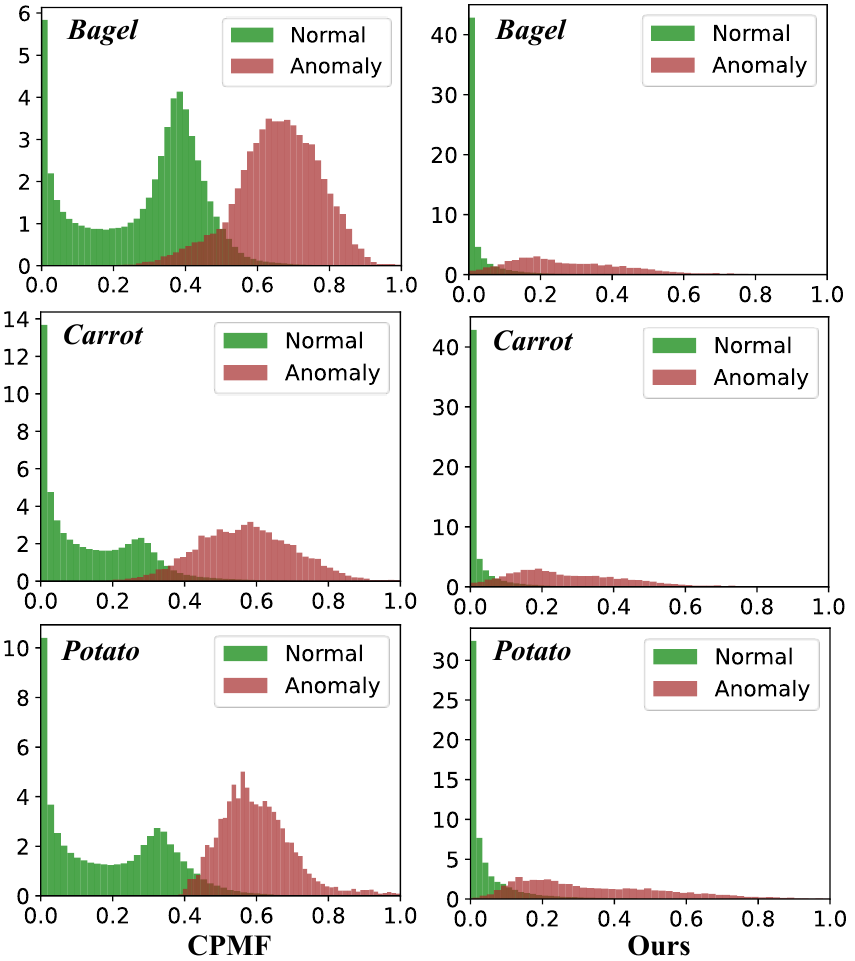}}
    \caption{Point-level anomaly score distributions across three categories. The x-axis represents anomaly scores, and the y-axis represents probability density. It can be observed that our approach effectively suppresses the anomaly scores in normal regions compared to CPMF.}
    \label{anomaly_dist}
\end{figure}

\textbf{Anomaly score distributions.}
Fig. \ref{anomaly_dist} illustrates the distribution of anomaly scores for CPMF and our method across three categories. It is evident that CPMF exhibits higher anomaly scores in normal regions, and the density of anomaly scores near zero is significantly lower than that of our method. For instance, in the \textit{Bagel} category, the CPMF has a density of 6, while our IAENet has a density of 40. In contrast, our approach effectively suppresses the anomaly scores of normal points, as evidenced by the anomaly scores for the majority of normal points being close to zero. This property enables our method to reduce the false positive rate, allowing it to outperform comparative methods even at lower integration limits (as shown in Fig. \ref{integration}), and thus has greater practical value in industrial applications.

\subsubsection{Comparison of Fusion Strategies}

To validate the effectiveness of the proposed IAF module, we compare it with existing score map fusion strategies, including pixel-wise addition \cite{horwitz2023back}, maximum selection \cite{chu2023shape}, and data-driven one-class Support Vector Machine (OCSVM) \cite{wang2023multimodal}. Table \ref{fusion comparison} presents the quantitative results of different fusion strategies, demonstrating that our IAF module consistently achieves the best performance across various metrics. This comparison underscores the superiority of our approach in effectively integrating the strengths of multiple models to enhance anomaly detection accuracy.

\begin{figure}[t]
    \centerline{\includegraphics[width = \linewidth]{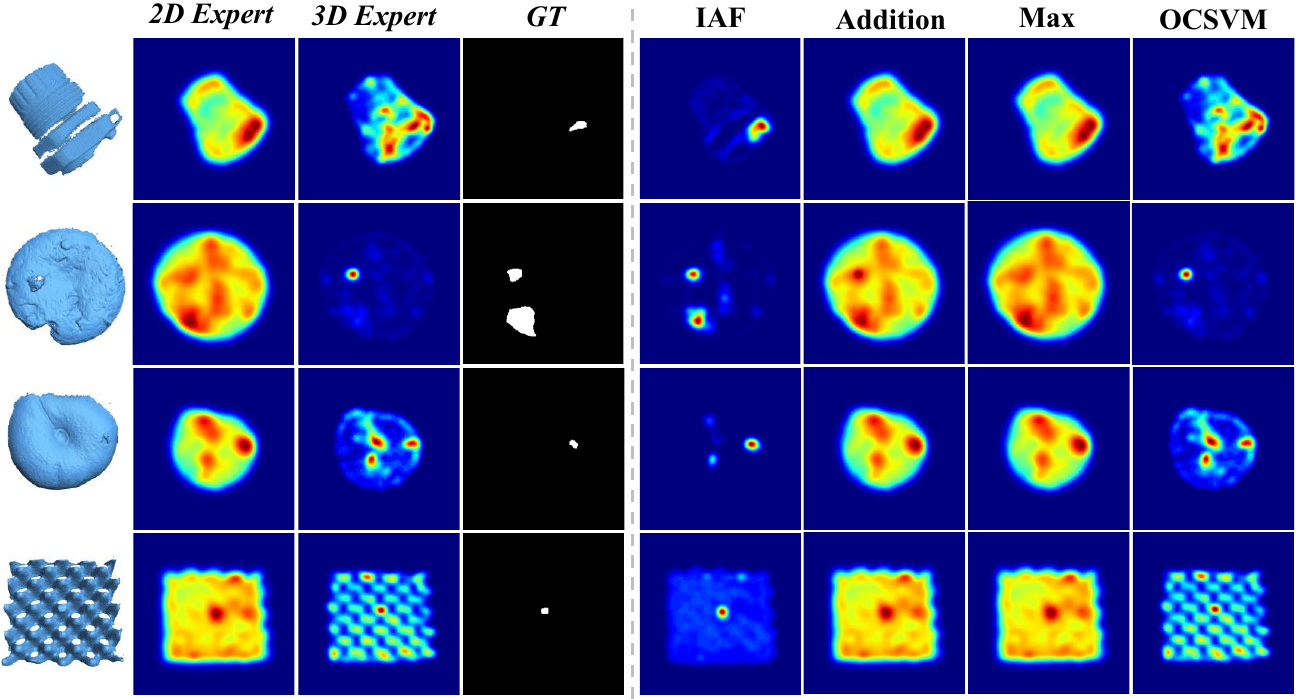}}
    \caption{Qualitative comparison of different fusion strategies in representative categories, including \textit{Cable gland, Cookie, Peach and Foam}. The left side shows the input point clouds, the fusion inputs and ground truth, and the right side shows the outputs of different fusion strategies.}
    \label{fusion_strategy}
\end{figure}

The visualization results for different fusion strategies are shown in Fig. \ref{fusion_strategy}. As we can see, the 3D expert tends to identify significant anomalies with high confidence, as indicated by the large difference between its normal and anomaly scores. Conversely, the 2D expert has a tendency to detect all potential anomalies, which is reflected in relatively high normal scores. Based on the outputs from the source experts, our IAF module not only integrates the collective knowledge of the source experts but also preserves their unique strengths. This means it can identify all potential anomalies with high confidence while suppressing scores in normal regions, thereby reducing the false positive rate. In contrast, other fusion strategies fail to achieve this balance. Addition and Max strategies are biased towards the 2D expert due to the higher values in its anomaly maps, while OCSVM favors the 3D expert, which performs better in the anomaly detection task. However, OCSVM cannot effectively utilize information from the 2D expert, potentially missing some anomalies, as illustrated in the \textit{Cookie} example in the Fig. \ref{fusion_strategy}.

\begin{table}[]
\centering
\caption{Mean values of different fusion strategies across all categories of the MVTec 3D-AD dataset.}
\resizebox{\linewidth}{!}{
\begin{tabular}{@{}lcccc@{}}
\toprule
Stategies        & O-AUROC        & P-AUROC        & AURPO@30\%     & AUPRO@1\% \\ \midrule
Max              & 0.921          & 0.974          & 0.916          & 0.367       \\
Addition         & 0.934          & 0.975          & 0.920          & 0.383       \\
OCSVM            & 0.921          & 0.977          & 0.929          & 0.399       \\
Ours             & \textbf{0.944} & \textbf{0.982} & \textbf{0.944} & \textbf{0.424}       \\ \bottomrule
\end{tabular}}
\label{fusion comparison}
\end{table}

\subsubsection{Ablation Study}

The IAF module is composed of two main components: the selector network $S^\theta$ and the predictor network $f^\phi_i$. The selector network is primarily responsible for evaluating the contributions of different experts to the final anomaly detection task and providing corresponding importance scores. The predictor network, on the other hand, performs a nonlinear mapping on the weighted score maps. The selector network is trained using a specially designed loss function, \( \mathcal{L}_{s} \). Therefore, in this section, we will examine the impact of the three critical components on the final anomaly detection performance: $S^\theta$, $f^\phi_i$, and \(\mathcal{L}_{s} \).

It can be observed in Table \ref{exp:abaltion} that when only the predictor network \( f^\phi \) is present, indicating a simple nonlinear mapping, the results are the poorest because it treats the outputs from different experts equally. Upon adding the selector network \( S^\theta \), the model attempts to assess the contributions of various experts, but is solely optimized by the end-to-end anomaly detection loss, which fails to yield reliable importance scores. However, with the inclusion of \( \mathcal{L}_{s} \), the selector network is explicitly guided in its optimization, enabling it to more accurately evaluate the contributions of different experts, thereby achieving the best anomaly detection results.  It is noteworthy that if only \( \mathcal{L}_{s} \) is used without $\mathcal{H}$, the model also fails to achieve satisfactory results, indicating that $\mathcal{H}$ plays a significant role in obtaining the optimal combination of weights for the final decision.

\begin{table}[]
\centering
\caption{Ablation study of our IAENet.}
\begin{tabular}{@{}lcccc@{}}
\toprule
\multicolumn{1}{c}{\textbf{Components}}                           & \multicolumn{4}{c}{\textbf{Stepwise Combinations}} \\  \midrule
Predictor Network $S^\theta$                                      & \checkmark   & \checkmark      & \checkmark      & \checkmark              \\
Selector Network $f^\phi_i$                                       &              & \checkmark      & \checkmark      & \checkmark              \\
$\mathcal{L}_{s}$ w/o $\mathcal{H}$                               &              &                 & \checkmark      & \checkmark              \\
$\mathcal{L}_{s}$ w/ $\mathcal{H}$                                &              &                 &                 & \checkmark              \\ \midrule
O-AUROC                                                           & 0.938  & 0.942  & 0.934  & \textbf{0.944} \\
P-AUROC                                                           & 0.966  & 0.970  & 0.971  & \textbf{0.982} \\
AURPO@30\%                                                        & 0.921  & 0.924  & 0.925  & \textbf{0.944} \\ \bottomrule 
\end{tabular}
\label{exp:abaltion}
\end{table}

\section{Conclusion}
\label{conclusion}
In this work we present IAENet, a novel ensemble framework that seamlessly integrates a 2D pretrained foundation model with a dedicated 3D expert to push the frontiers of 3D point cloud-based anomaly detection. Recognizing that naive fusion of heterogeneous predictions can be undermined by large inter-expert discrepancies, we introduce the Importance-Aware Fusion (IAF) module together with specifically designed loss functions. IAF dynamically reweights the contribution of each model, ensuring that the final decision leverages their complementary strengths while preserving their individual merits.
Extensive experiments on the MVTec 3D-AD benchmark confirm that IAENet establishes a new state of the art. Ablation studies further validate the necessity of every component within IAF. Notably, our method effectively suppresses anomaly scores on normal regions, yielding markedly lower false positive rates, an attribute of critical value for real-world industrial deployment.

\printcredits

\section*{Acknowledgments}
This work is supported by the National Natural Science Foundation of China (No. 72371219 and No. 72001139), and the Guangzhou Municipal Science and Technology Project (No. 2025A04J5288).

\bibliographystyle{cas-model2-names}

\bibliography{cas-refs}



\end{document}